\documentclass[a4paper]{llncs}

\usepackage{amssymb}
\usepackage{graphicx}
\usepackage[utf8]{inputenc}

\usepackage{url} 
\urldef{\mailsc}\path|{collinsl, tmarwala}@uj.ac.za|  
\newcommand{\keywords}[1]{\par\addvspace\baselineskip
\noindent\keywordname\enspace\ignorespaces#1}

\begin{document}

\mainmatter  

\title{Missing Data Estimation in High-Dimensional Datasets: A Swarm Intelligence-Deep Neural Network Approach}


%
%
\author{Collins Leke\and Tshilidzi Marwala}
\institute{University of Johannesburg, Johannesburg, South Africa
\mailsc}

%
%

\maketitle

\begin{abstract}
In this paper, we examine the problem of missing data in high-dimensional datasets by taking into consideration the Missing Completely at Random and Missing at Random mechanisms, as well as the Arbitrary missing pattern. Additionally, this paper employs a methodology based on Deep Learning and Swarm Intelligence algorithms in order to provide reliable estimates for missing data. The deep learning technique is used to extract features from the input data via an unsupervised learning approach by modeling the data distribution based on the input. This deep learning technique is then used as part of the objective function for the swarm intelligence technique in order to estimate the missing data after a supervised fine-tuning phase by minimizing an error function based on the interrelationship and correlation between features in the dataset. The investigated methodology in this paper therefore has longer running times, however, the promising potential outcomes justify the trade-off. Also, basic knowledge of statistics is presumed.
\end{abstract}

\section{Introduction}
Previous research across a wide range of academic fields suggests that decision-making and data analysis tasks are made nontrivial by the presence of missing data. As such, it can be assumed that decisions are likely to be more accurate and reliable when complete/representative datasets are used instead of incomplete datasets. This assumption has led to a lot of research in the data mining domain, with novel techniques being developed to perform this task accurately \cite{abdella2005use}-\cite{zhang2011shell}. Research suggests that applications in various professional fields such as in medicine, manufacturing or energy that use sensors in instruments to report vital information and enable decision-making processes may fail and lead to incorrect outcomes due to the presence of missing data. In such cases, it is very important to have a system capable of imputing the missing data from the failed sensors with high accuracy. The imputation procedure will require the approximation of missing values taking into account the interrelationships that exist between the data from sensors in the system. Another instance where the presence of missing data poses a threat in decision-making is in image recognition systems, whereby the absence of pixel values renders the image prediction or classification task difficult and as such, systems capable of imputing the missing values with high accuracy are needed to make the task more feasible.
\begin{figure}
	\includegraphics[width = \textwidth, height = 0.7in]{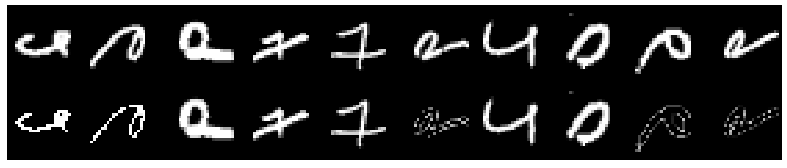}
	\caption{Sample of MNIST Dataset. Top Row - Real Data: Bottom Row - Data With Missing Pixel Values}
	\label{fig:missingpixels}
\end{figure}

Consider a high dimensional dataset such as the Mixed National Institute of Standards and Technology (MNIST) dataset with 784 feature variables being the pixel values as shown in Fig. \ref{fig:missingpixels} above. Assuming that pixel values are missing at random as observed in the bottom row and a statistic analysis is required to classify the above dataset, the questions of interest would be: (i) Can we impute with some degree of certainty the missing data in high dimensional datasets with high accuracy? (ii) Can new techniques be introduced for approximation of the missing data when correlation and interrelationships between the variables are considered? This paper therefore aims to use a Deep Learning (DL) technique built with Restricted Boltzmann machines stacked together to form an autoencoder in tandem with a swarm intelligence (SI) algorithm to estimate the missing data with the model created which would cater to the mechanisms of interest and the arbitrary pattern. The dataset used is the MNIST database of handwritten digits by Yann LeCun \cite{LeCun}. It has a training set of 60,000 sample images and a test set of 10,000 sample images with 784 features. These images show handwritten digits from 0 to 9. Due to the fact that the research discussed in this paper was conducted at a time when there was little or no interest in the DL-SI missing data predictors on high dimensional data, this paper seeks to exploit the use of this technique on the MNIST dataset. 

The remainder of this paper is structured as follows, Section 2 introduces missing data, the deep learning techniques used as well as the swarm intelligence algorithm implemented. This section also presents related work in the domain. Section 3 presents the experimental design and procedures used, while Section 4 focuses on the results and key findings from the experiments conducted in this article. Discussions, concluding remarks and suggestions for future research are further presented in Section 5.

\section{Background}
This article implements a Deep Learning technique referred to as a Stacked Autoencoder built using Restricted Boltzmann machines, all of which have been individually trained using the Contrastive Divergence algorithm and stacked together in a bottom-up manner. The estimation of missing values is performed by using the Firefly Algorithm, which is the swarm intelligence method. However, this article will first briefly discuss the methods used and the problem it aims to solve.
\subsection{Missing Data and Deep Learning} 
Missing data is a situation whereby some features within a dataset are lacking components \cite{rubin1978multiple}. With this ensues problems in application domains that rely on the access to complete and quality data which can affect every academic/professional fields and sectors. Techniques aimed at rectifying the problem have been an area of research in several disciplines \cite{rubin1978multiple}-\cite{little2014statistical}. The manner in which data points go missing in a dataset determines the approach to be used in estimating these values. As per \cite{little2014statistical}, there exist three missing data mechanisms. This article focuses on investigating the Missing Completely at Random (MCAR) and Missing at Random (MAR) mechanisms. Previous research suggests that MCAR scenario arises when the chances of there being a missing data entry for a feature is not dependent on the feature itself or on any of the other features in the dataset \cite{leke2014modeling}. This implies a lack of correlation or cross-correlation between features including the feature of interest \cite{rubin1978multiple}. MAR on the other hand arises when missingness in a specific feature is reliant upon the other features within the dataset, but not the feature of interest itself \cite{leke2014modeling}. According to \cite{little2014statistical}, there are two main missing data patterns. These are the arbitrary and monotone missing data patterns. In the arbitrary pattern, missing observations may occur anywhere and the ordering of the variables is of no importance. In monotone missing patterns, the ordering of the variables is of importance and occurrence is not random. Based upon this realization, this article will go on to focus on the arbitrary missing pattern.

Deep Learning comprises of several algorithms in machine learning that make use of a cataract of nonlinear processing units organized into a number of layers that extract and transform features from the input data \cite{deng2013recent}, \cite{deng2014deep}. Each of the layers use the output from the previous layer as input and a supervised or unsupervised algorithm could be used in the training phase. With these come applications in supervised and unsupervised problems like classification and pattern analysis, respectively. It is also based on the unsupervised learning of multiple levels of features or representations of the input data whereby higher-level features are obtained from lower level features to yield a hierarchical representation of the data \cite{deng2014deep}. By learning multiple levels of representations that depict different levels of abstraction of the data, we obtain a hierarchy of concepts. In this article, the Deep Learning technique used is the Stacked AutoEncoder.

\subsection{Restricted Boltzmann Machine (RBM)}\label{RBM}
Firstly, a Boltzmann machine (BM) is an undirected network with nodes possessing stochastic traits that can be described as a neural network. It is used amongst other things to extract vital information from an unknown probability distribution using samples from the distribution, which is generally a difficult process \cite{Fischer2012}. This learning process is made simple by implementing restrictions on the network structure leading to Restricted Boltzmann machines (RBMs). An RBM can be described as an undirected, probabilistic, parameterized graphical model also known as a Markov random field (MRF). RBMs became techniques of interest after being suggested as components of multi-layer topologies termed deep networks \cite{Fischer2012}. The idea is that hidden nodes extract vital information from the observations, which subsequently represent inputs to the next RBM. Stacking these RBMs together has as objective, obtaining high level representations of data by learning features from features. An RBM which is also an MRF associated with a bipartite undirected graph consists of $m$ visible nodes, $V = (V_{1},...,V_{m})$ representing input data, and $n$ hidden nodes, $H = (H_{1},...,H_{n})$ capturing interdependencies between features in the input layer \cite{Fischer2012}. In this article, the features $V$ have as values, $v \in [0,1]^{m+n}$, while $H$ have as values, $h \in \left\lbrace0,1\right\rbrace^{m+n}$. The distribution given by the Gibbs distribution has as energy function \cite{Fischer2012}:
\begin{equation}
E\left(v,h\right) = -h^{T}Wv - b^{T}v - c^{T}h \ .
\label{MatrixE}
\end{equation}
In scalar form, (\ref{MatrixE}) is expressed as \cite{Fischer2012}:
\begin{equation}
E\left(v,h\right) = -\sum_{i=1}^{n}\sum_{j=1}^{m}w_{ij}h_{i}v_{j} - \sum_{j=1}^{m}b_{j}v_{j} - \sum_{i=1}^{n}c_{i}h_{i} \ .
\label{ScalarE}
\end{equation}
In (\ref{ScalarE}), $w_{ij}$, which is the most important part of an RBM model is a real valued weight between units $V_{j}$ and $H_{i}$, while $b$ and $c$ are the bias terms for the visible and hidden variables, respectively. If $w_{ij}$ is negative, and $v_{j}$ and $h_{i}$ are equal to one, the probability decreases leading to a high energy. On the contrary, if $w_{ij}$ is positive, and $v_{j}$ and $h_{i}$ are equal to zero, the probability increases leading to a lower energy. If $b_{j}$ is negative and $v_{j} = 1$, $E$ increases leading to a low probability. Therefore, there is a preference for $v_{j} = 0$ instead of $v_{j} = 1$. However, if $b_{j}$ is positive and $v_{j} = 0$, $E$ decreases leading to a high probability, and a preference for $v_{j} = 1$ instead of $v_{j} = 0$. A negative $b_{j}$ value decreases the second term in (\ref{ScalarE}), while a positive value for $b_{j}$ increases this second term. The same applies for $c_{i}$ and the third term in (\ref{ScalarE}). The Gibbs distributions or probabilities from (\ref{MatrixE}) or (\ref{ScalarE}) are then obtained by \cite{Fischer2012}:
\begin{equation}
p\left(v,h\right) = \frac{e^{-E\left(v,h\right)}}{Z} = \frac{e^{(h^{T}Wv + b^{T}v + c^{T}h)}}{Z} = \frac{e^{(h^{T}Wv)}e^{(b^{T}v)}e^{(c^{T}h)}}{Z} \ .
\label{Dist}
\end{equation}
Here, the exponential terms are factors of a markov network with vector nodes, while $Z$ is the intractable partition function. It is intractable courtesy of the exponential number of values it can take. For an RBM, $Z = \sum_{v,h}e^{-E\left(v,h\right)}$. Another key aspect of RBMs is that $h$ is conditionally independent of $v$ and vice versa, due to the fact that there are no connections between nodes in the same layer. This property is expressed mathematically as \cite{Fischer2012}:
\begin{equation}
p(h|v) = \prod_{i=1}^{n}p(h_{i}|v) \ and \ p(v|h) = \prod_{i=1}^{m}p(v_{i}|h) \ .
\label{conditional}
\end{equation}

\subsection{Contrastive Divergence (CD)}\label{CD}
The objective in training an RBM is to minimize the average negative log-likelihood (loss) without regularization using a stochastic gradient descent algorithm as it scales well with high-dimensional datasets. Achieving this objective requires the partial derivative of any parameter, $\theta$, of the loss function as per the following equation:
\begin{equation}\label{eq4}
\frac{\partial \left(-logp\left(v^{\left(t\right)}\right)\right)}{\partial\theta} = E_{h}\left[\frac{\partial E\left(v^{\left(t\right)},h\right)}{\partial\theta} | v^{\left(t\right)}\right] - E_{v,h}\left[\frac{\partial E\left(v,h\right)}{\partial\theta}\right] \ .
\end{equation}
The first term in (\ref{eq4}) is the expectation over the data distribution and is referred to as the positive phase, while $v$ and $h$ represent the same variables as in (\ref{MatrixE})-(\ref{conditional}). The second term, which is the expectation over the model distribution is termed the negative phase. This phase is hard to compute and also intractable because an exponential sum is required over both $h$ and $v$. Furthermore, many sampling steps are needed to obtain unbiased estimates of the log-likelihood gradient. However, it has been shown recently that running a markov chain for just a few steps leads to estimates that are sufficient for training a model \cite{Fischer2012}. This approach has led to the contrastive divergence (CD) algorithm. CD is a training method for undirected probabilistic graphical models with the idea being to do away with the double expectations in the negative phase in (\ref{eq4}) and instead focus on estimation. It basically implements a Monte-Carlo estimate of the expectation over a single input data point. The idea of $k$-step CD (CD-$k$) is that rather than the second term being approximated in (\ref{eq4}) by a sample from the model distribution, $k$ steps of a Gibbs chain is run, with $k$ frequently set to 1 \cite{Fischer2012}. The Gibbs chain starts with a training sample $v^{(0)}$ of the training data and returns $v^{(k)}$ after $k$ steps \cite{Fischer2012}. Each step, $t$, entails sampling $h^{(t)}$ from $p(h|v^{(t)})$, then obtaining samples $v^{(t+1)}$ from $p(v|h^{(t)})$ \cite{Fischer2012}. For one training pattern, $v^{(0)}$, the log-likelihood gradient w.r.t. $\theta$ is approximated by \cite{Fischer2012}:
\begin{equation}\label{app}
CD_{k}(\theta,v^{(0)}) = - \sum_{h}p(h|v^{(0)})\frac{\partial E(v^{(0)},h)}{\partial \theta} + \sum_{h}p(h|v^{(k)})\frac{\partial E(v^{(k)},h)}{\partial \theta} \ .
\end{equation}
Due to the fact that $v^{(k)}$ is not a obtained from the stationary model distribution, the approximation (\ref{app}) is biased. The bias in effect fades away as $k \longrightarrow \infty$ \cite{Fischer2012}. Another aspect that points to CD being biased is that it maximizes the difference between two Kullback-Liebler (KL) divergences \cite{Fischer2012}:
\begin{equation}
KL(q|p) - KL(p_{k}|p) \ .
\end{equation} 
Here, the experimental distribution is $q$ and the distribution of the visible variables after $k$ steps of the Markov chain is $p_{k}$ \cite{Fischer2012}. If stationarity in the execution of the chain is already attained, $p_{k} = p$ holds, and therefore $KL(p_{k}|p) = 0$, and the error of the approximation by CD fades away \cite{Fischer2012}.

\subsection{Autoencoder (AE)}\label{AE}
An Autoencoder is an artificial neural network that attempts to reproduce its input at the output layer. The basic idea behind autoencoders is that the mapping from the input to the output, $ x^{\left(i\right)} \mapsto y^{\left(i\right)}$ reveals vital information and the essential structure in the input vector $x^{\left(i\right)}$ that is otherwise abstract. An autoencoder takes an input vector $x$ and maps it to a hidden representation $y$ via a deterministic mapping function $f_{\theta}$ of the form $f_{\theta} \left(x\right) = s\left(Wx + b\right)$ \cite{isaacs2014}. The $\theta$ parameter comprises of the matrix of weights $W$ and the vector of offsets/biases $b$. $s$ is the sigmoid activation function expressed as: 
\begin{equation}
s = \frac{1}{1+e^{-x}} \ . 
\end{equation}
The hidden representation $y$ is then mapped to a reconstructed vector $z$ which is obtained by the functions \cite{Fischer2012}:
\begin{equation}
z = g_{\theta^{'}}\left(y\right) = s\left(W'y + b'\right) \ or \ z = g_{\theta^{'}}\left(y\right) = W'y + b' \ .
\end{equation}
Here, the parameter set $\theta^{'}$ comprises of the transpose of the matrix of weights and vector of biases from the encoder prior to the fine-tuning phase \cite{isaacs2014}. When the aforementioned transposition of weights is done, the autoencoder is said to have tied weights. $z$ is not explained as a rigorous regeneration of $x$ but instead as the parameters of $p\left(X|Z = z\right)$ in probabilistic terms, which may yield $x$ with high probability \cite{isaacs2014}. This thus leads to:
\begin{equation}
p\left(X|Y = y\right) = p\left(X|Z = g_{\theta^{'}}\left(y\right)\right) \ .
\end{equation} 
From this, we obtain a reconstruction error which is to be optimized by the optimization technique and is of the form $L\left(x, z\right) \propto -logp\left(x|z\right)$. This equation as per \cite{bengio2013representation} could also be expressed as:
\begin{equation}
\delta_{AE}\left(\theta\right) = \sum_{t}L\left(x^{\left(t\right)},g_{\theta}\left(f_{\theta}\left(x^{\left(t\right)}\right)\right)\right) \ .
\end{equation}  

\subsection{Firefly Algorithm (FA)}\label{FA}
FA is a nature-inspired metaheuristic algorithm based on the flashing patterns and behavior of fireflies \cite{xin}. It is based on three main rules being: (i) Fireflies are unisex so all fireflies are attracted to all other fireflies, (ii) Attractiveness is proportional to the brightness and they both decrease as the distance increases. The idea is the less brighter firefly will move towards the brighter one. If there is no obvious brighter firefly, they move randomly, and, (iii) Brightness of a firefly is determined by the landscape of the objective function \cite{xin}. Considering that attractiveness is proportional to light intensity, the variation of attractiveness can be defined with respect to the distance as \cite{xin}:
\begin{equation}\label{attract}
\beta = \beta_{0}e^{-\gamma r^{2}} \ .
\end{equation}
In (\ref{attract}), $\beta$ is the attractiveness of a firefly, $\beta_{0}$ is the initial attractiveness of a firefly, and $r$ is the distance between two fireflies. The movement of a firefly towards a brighter one is determined by \cite{xin}:
\begin{equation}\label{move}
x_{i}^{t+1} = x_{i}^{t} + \beta_{0}e^{-\gamma r_{ij}^{2}}\left(x_{j}^{t} - x_{i}^{t}\right) + \alpha_{t}\epsilon_{i}^{t} \ .
\end{equation}
Here, $x_{i}$ and $x_{j}$ are the positions of two fireflies, and the second term is due to the attraction between the fireflies. $t$ and $t+1$ represent different time steps, $\alpha$ is the randomization parameter controlling the step size in the third term, while $\epsilon$ is a vector with random numbers obtained from a Gaussian distribution. If $\beta_{0} = 0$, the movement is then a simple random walk \cite{xin}. If $\gamma = 0$, the movement reduces to a variant of the particle swarm optimization algorithm \cite{xin}. The parameters used in this research are: (i) n = number of missing cases per sample, (ii) 1000 iterations, (iii) $\alpha = 0.25$, (iv) $\beta = 0.2$ and (v) $\gamma = 1$. The parameters were selected as they yielded the more optimal results after experimemtation with different permutations and combinations of values. The FA algorithm is used because although it has been successfully applied in a number of domains such as digital image compression, eigenvalue optimization, feature selection and fault detection, scheduling and TSP, etc., its efficiency has not been investigated in missing data estimation tasks on high-dimensional datasets.

\subsection{Related Work}\label{Related Work}
We present some of the work that has been done by researchers to address the problem of missing data. The research done in \cite{abdella2005use} implements a hybrid genetic algorithm-neural network system to perform missing data imputation tasks with varying number of missing values within a single instance while \cite{aydilek2012novel} creates a hybrid k-Nearest Neighbor-Neural Network system for the same purpose. In \cite{leke2014modeling}, a hybrid Auto-Associative neural network or autoencoder with genetic algorithm, simulated annealing and particle swarm optimization model is used to impute missing data with high levels of accuracy in cases where just one feature variable has missing input entries. In some cases, neural networks were used with Principal Component Analysis (PCA) and genetic algorithm as in \cite{mistry2009missing}-\cite{nelwamondo2007missing}. In \cite{rana2015robust}, they use robust regression imputation for missing data in the presence of outliers and investigate its effectiveness. In \cite{zhang2011missing}, it is suggested that information within incomplete cases, that is, instances with missing values be used when estimating missing values. A nonparametric iterative imputation algorithm (NIIA) is proposed that leads to a root mean squared error value of at least 0.5 on the imputation of continuous values and a classification accuracy of at most 87.3\% on the imputation of discrete values with varying ratios of missingness. In \cite{zhang2011shell}, the shell-neighbor method is applied in missing data imputation by means of the Shell-Neighbor Imputation (SNI) algorithm which is observed to perform better than the k-Nearest Neighbor imputation method in terms of imputation and classification accuracy as it takes into account the left and right nearest neighbors of the missing data as well as varying number of nearest neighbors contrary to k-NN that considers just fixed \textit{k} nearest neighbors. In \cite{lobato2015multi}, a multi-objective genetic algorithm approach is presented for missing data imputation. It is observed that the results obtained outperform some of the well known missing data methods with accuracies in the 90 percentile. Novel algorithms for missing data imputation and comparisons between existing techniques can be found in papers such as \cite{lobato2015multi}-\cite{van2012flexible}.

\section{Experimental Design and Procedure}
In the design of the experiments, MATLAB R2014a software was used on a Dell Desktop computer with Intel(R) Core(TM) i3-2120 CPU @ 3.30GHz processor, 4.00 GB RAM, 32 GB virtual RAM, 64-bit Operating System running Windows 8.1 Pro. Additionally, the MNIST database was used and it contains 60,000 training images and 10,000 test images. Each of these images is of size $28\times28 = 784$ pixels. This results in a training set of size $60000\times784$ and a test of size $10000\times784$. Data preprocessing was performed normalizing all pixel values in the range [0, 1]. The individual network layers of the Deep AE were pretrained using RBMs and CD to initialize the weights and biases in a good solution space. The individual layers pretrained were of size $784-1000$, $1000-500$, $500-250$, and $250-30$. These are stacked and subsequently transposed to obtain the encoder and decoder parts of the autoncoder network, respectively. The resulting network architecture is of size, $784-1000-500-250-30-250-500-1000-784$, with an input and output layer with the same number of nodes, and seven hidden layers with varying number of nodes. The network is then fine-tuned using backpropagation, minimizing the mean squared network error. The error value obtained after training is 0.0025. The training is done using the entire training set of data that are divided into 600 balanced mini-batches. The weight and bias updates are done after every mini-batch. Training higher layers of weights is achieved by having the real-valued activations of the visible nodes in preceeding RBMs being transcribed as the activation probabilities of the hidden nodes in lower level RBMs. The Multilayer Perceptron (MLP) AE has an input and output layer, both consisting of 784 nodes, and one hidden layer consisting of 400 nodes obtained by experimenting with different numbers of nodes in the hidden layer, and observing which architecture leads to the lowest mean squared network error. A $784-400-784$ network architecture led to the lowest mean squared network error value of 0.0032. The hidden and output layer activation function used is the sigmoid funtion. The training is done using the scaled conjugate gradient descent algorithm for 1000 epochs. Missingness in the test set of data is then created at random according to the MAR and MCAR mechanisms, as well as the arbitrary pattern, and these missing values are approximated using the swarm intelligence algorithm which has as objective function minimizing the loss function of the fine-tuned network. The tolerance error is intially set to 0.05 (5\% ) in one of the networks, and is considered reasonable for a first time investigation of the proposed method. The overall approach consist of four consecutive steps being:
\begin{enumerate}
	\item Train the individual RBMs on a training set of data with complete records using the greedy layer-by-layer pre-training algorithm described in \cite{hinton2006reducing} starting from the bottom layer. Each layer is trained for 50 epochs with the learning rate for the weights, visible unit biases and hidden unit biases set to 0.1. The initial and final momentum are set to 0.5 and 0.9, respectively. The final parameter is the weight cost which is set to 0.0002.
	\item Stack the RBMs to form the Encoder and Decoder phases of a Deep Autoencoder with tied weights.
	\item Fine-tune the Deep Autoencoder using back-propagation for 1000 epochs through the entire set of training data.
	\item Estimate the missing data with the fine-tuned deep network as part of the objective function in the Firefly Algorithm parsing the known variable values to the objective function, while first estimating the unknown values before parsing these estimates to the objective function. The estimation procedure is terminated when a stopping criterion is achieved, which is either an error tolerance of 5\% (0.05), or the maximum number of function evaluations being attained.
\end{enumerate}

\section{Experimental Results}
\begin{figure}\centering
	\begin{minipage}{0.49\textwidth}
		\frame{\includegraphics[width=\linewidth]{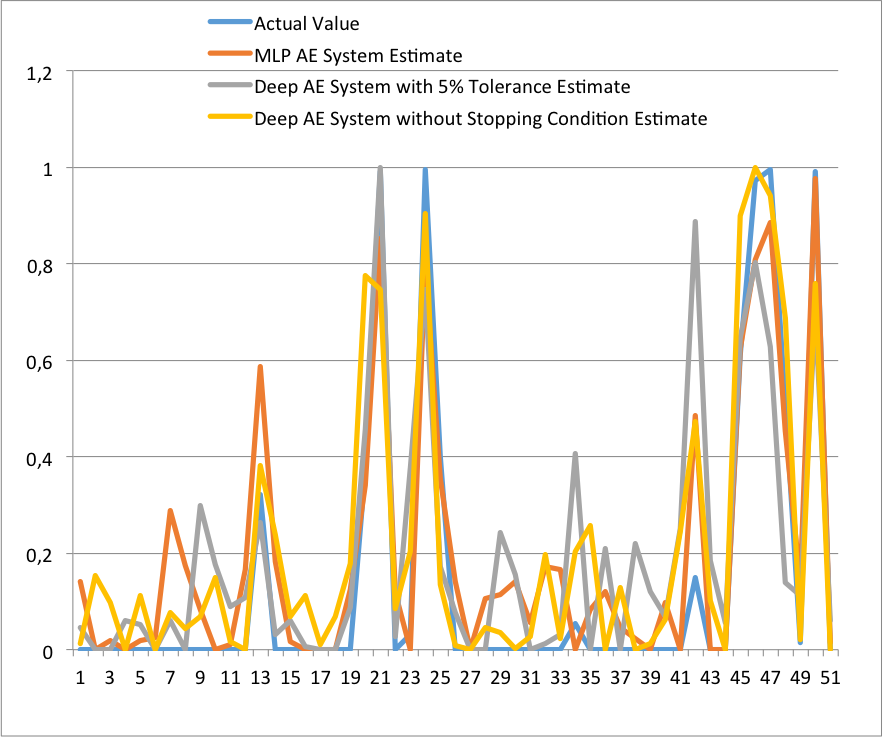}}
		\caption{Actual vs Estimated Values.}\label{Fig:fig2}
	\end{minipage}
	\begin{minipage}{0.49\textwidth}
		\frame{\includegraphics[width=\textwidth]{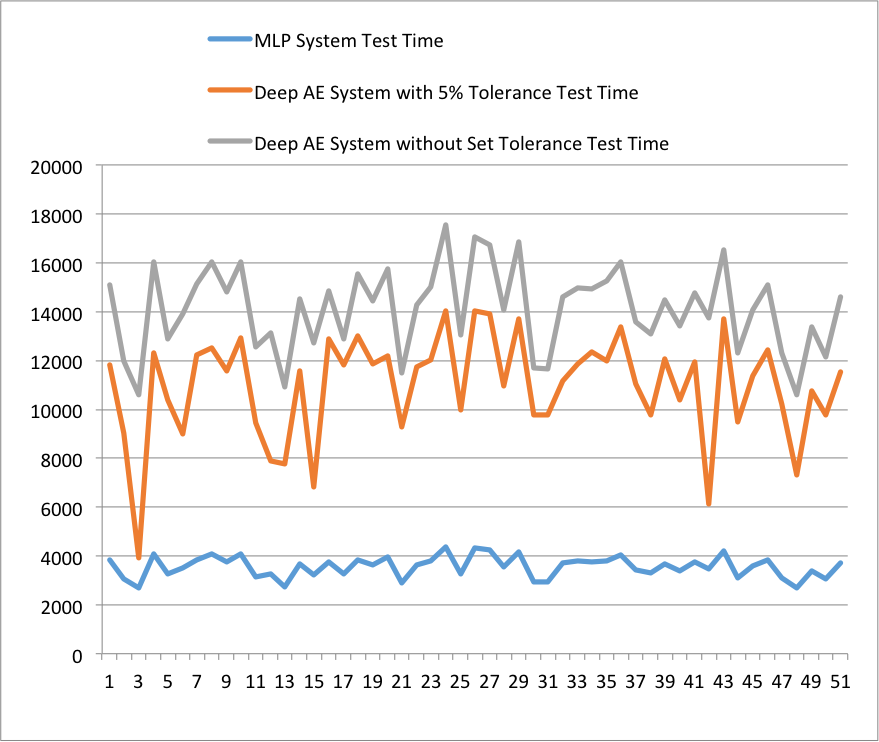}}
		\caption{Test Times per Sample.}\label{Fig:fig3}
	\end{minipage}
\end{figure}

In the investiagtion of the imputation technique, we used the test set of data which contained missing data entries accounting for approximately 10\% of the data. We present in Tables \ref{tab:table1} and \ref{tab:table2}, Actual, Estimate and Squared Error values from the proposed Deep Autoencoder system without tolerance (Table \ref{tab:table1}), and from MLP Autoencoder system (Table \ref{tab:table2}). The distance, $\epsilon$, from the estimate to the actual value, added to the squared error are parameters that determine the performance of the method. In all cases presented in both tables, the Deep Autoencoder system shows $\epsilon_{d} = 0, 0.0608, 0, 0.0275, 0, 0.0922, 0.0009, 0.0283$, while for the same entries (actual values), the MLP Autoencoder shows that $\epsilon_{m} = 0.0246, 0.2646, 0.0149, 0.1643, 0, 0.1982, 0.0509, 0.0473$, respectively. They show better performance of the proposed technique without a set error tolerance when compared to the existing MLP Autoencoder. This knowledge is validated by the squared error which is always smaller for the proposed technique, for all cases presented in Tables \ref{tab:table1} and \ref{tab:table2}. We could consider this enough to conclude of on the performance of both compared techniques, but we need to analyse the processing time, which seems to be better for the existing method when compared to the proposed Deep Autoencoder system. This is demonstrated by Fig. \ref{Fig:fig3}, where we compare processing times for both techniques. It is evident that setting an error tolerance value makes the estimation process faster as observed in Fig. \ref{Fig:fig3}. However, this is at the expense of accuracy which is the main aspect in such a task as seen in Fig. \ref{Fig:fig2}. The bigger the error tolerance value, the faster the estimation of the missing data.
\begin{table}
	\centering
	\setlength\tabcolsep{4pt}
	\begin{minipage}{0.49\textwidth}
		\centering
		\caption{Actual, Estimated and Squared Error Values from Deep Autoencoder System without Set Tolerance.}
		\begin{tabular}{|c|c|c|}\hline
			Actual & Estimate & Squared Error \\ \hline \hline 
			0 & 0 & 0 \\ \hline
			0.3216 & 0.3824 & 0.0037 \\ \hline
			0 & 0 & 0 \\ \hline
			0.9725 & 1 & 0.0008 \\ \hline
			0 & 0 & 0 \\ \hline
			0.9961 & 0.9039 & 0.0085 \\ \hline
			0.0509 & 0.0500 & 8.38e-07 \\ \hline
			0.5765 & 0.6048 & 0.0008 \\ \hline
		\end{tabular}
		\label{tab:table1} 
	\end{minipage}%
	\hfill
	\begin{minipage}{0.49\textwidth}
		\centering
		\caption{Actual, Estimated and Squared Error Values from MLP Autoencoder System without Set Tolerance.}
		\begin{tabular}{|c|c|c|}\hline
			Actual & Estimate & Squared Error \\ \hline \hline 
			0 & 0.0246 & 0.0006 \\ \hline
			0.3216 & 0.5862 & 0.0700 \\ \hline
			0 & 0.0149 & 0.0002 \\ \hline
			0.9725 & 0.8082 & 0.0270 \\ \hline
			0 & 0 & 0 \\ \hline
			0.9961 & 0.7979 & 0.0393 \\ \hline
			0.0509 & 0 & 0.0026 \\ \hline
			0.5765 & 0.5292 & 0.0022 \\ \hline
		\end{tabular} 
		\label{tab:table2} 
	\end{minipage}
\end{table}

\section{Discussion and Conclusion}
This paper investigates the use of a deep neural network with a swarm intelligence algorithm to impute missing data in a high-dimensional dataset. According to the arbitrary missing data pattern, MAR and MCAR mechanisms, missing data could occur anywhere in the dataset. The experiment in this paper considers a scenario in which 10\% of the test set of data is missing. These values are to be estimated with a set error tolerance of 5\%, as well as no set error tolerance. Also, the proposed method is compared to an MLP Autoencoder estimation system. The results obtained reveal that the proposed system yields the more accurate estimates, especially when there is no set error tolerance value. This is made evident when the distance and squared error values are considered. The AE systems both yield better estimates than the MLP system. However, with these accurate estimates come longer running times which are observed to become smaller when error tolerance values are set. The bigger the tolerance value, the smaller the running time. Based on the findings in this article, we intend to perform an in-depth parameter analysis in any future research in order to observe which parameters are optimal for the task and we will generalize this aspect using several datasets. Another obstacle faced in this research was the computation time to estimate the missing values and to address this, we will parallelize the process on a multi-core system to observe whether parallelizing the task does indeed speed up the process and maintain efficiency.

\end{document}